\documentclass[10pt,twocolumn,letterpaper]{article}

\usepackage{iccv}
\usepackage{times}
\usepackage{epsfig}
\usepackage{graphicx}
\usepackage{amsmath}
\usepackage{amssymb}

\usepackage[pagebackref=true,breaklinks=true,letterpaper=true,colorlinks,bookmarks=false]{hyperref}

\usepackage{booktabs}
\usepackage{subcaption}
\usepackage{multirow}
\usepackage{mmstyle}
\usepackage{mathtools}
\usepackage{amsmath}
\usepackage{hanging}
\usepackage{tablefootnote}
\usepackage[title]{appendix}

\usepackage[flushmargin]{footmisc}

\newcommand{\algname}{CARAFE}

\iccvfinalcopy 

\ificcvfinal\fi
\begin{document}

\title{CARAFE: Content-Aware ReAssembly of FEatures}

\author{Jiaqi Wang$^{1}$ \quad Kai Chen$^{1}$ \quad Rui Xu$^1$ \quad Ziwei Liu$^1$ \quad Chen Change Loy$^2$ \quad Dahua Lin$^1$\\
	$^1$CUHK - SenseTime Joint Lab, The Chinese University of Hong Kong \\
	$^2$Nanyang Technological University\\
	{\tt\small \{wj017,ck015,xr018,dhlin\}@ie.cuhk.edu.hk}\hspace{10pt}
	{\tt\small zwliu.hust@gmail.com}\hspace{10pt}
	{\tt\small ccloy@ntu.edu.sg}
}
\maketitle
\thispagestyle{empty}


\begin{abstract}

Feature upsampling is a key operation in a number of modern convolutional network architectures, \eg~feature pyramids. 
Its design is critical for dense prediction tasks such as object detection and semantic/instance segmentation.
In this work, we propose Content-Aware ReAssembly of FEatures (CARAFE), a universal, lightweight and highly effective operator to fulfill this goal. 
CARAFE has several appealing properties:
(1) Large field of view. Unlike previous works (\eg bilinear interpolation) that only exploit sub-pixel neighborhood, CARAFE can aggregate contextual information within a large receptive field.
(2) Content-aware handling. Instead of using a fixed kernel for all samples (\eg deconvolution), CARAFE enables instance-specific content-aware handling, which generates adaptive kernels on-the-fly.
(3) Lightweight and fast to compute. CARAFE introduces little computational overhead and can be readily integrated into modern network architectures. 
We conduct comprehensive evaluations on standard benchmarks in object detection, instance/semantic segmentation and inpainting.
CARAFE shows consistent and substantial gains across all the tasks (1.2\% AP, 1.3\% AP, 1.8\% mIoU, 1.1dB respectively) with negligible computational overhead.
It has great potential to serve as a strong building block for future research. Code and models are available at \url{https://github.com/open-mmlab/mmdetection}.

\end{abstract}


\section{Introduction}
\label{sec:intro}

Feature upsampling is one of the most fundamental operations in deep neural networks.
On the one hand, for the decoders in dense prediction tasks (\eg super resolution~\cite{Dong_2016, Lim_2017}, inpainting~\cite{iizuka2017globally, pathak2016context} and semantic segmentation~\cite{zhao2017pyramid, Chen_2018}), the high-level/low-res feature map is upsampled to match the high-resolution supervision. 
On the other hand, feature upsampling is also involved in fusing a high-level/low-res feature map with a low-level/high-res feature map, which is widely adopted in many state-of-the-art architectures, \eg, Feature Pyramid Network~\cite{lin2017feature}, U-Net~\cite{ronneberger2015u} and Stacked Hourglass~\cite{Newell_2016}.
Therefore, designing effective feature upsampling operator becomes a critical issue.

\begin{figure}
	\centering
	\includegraphics[width=\linewidth]{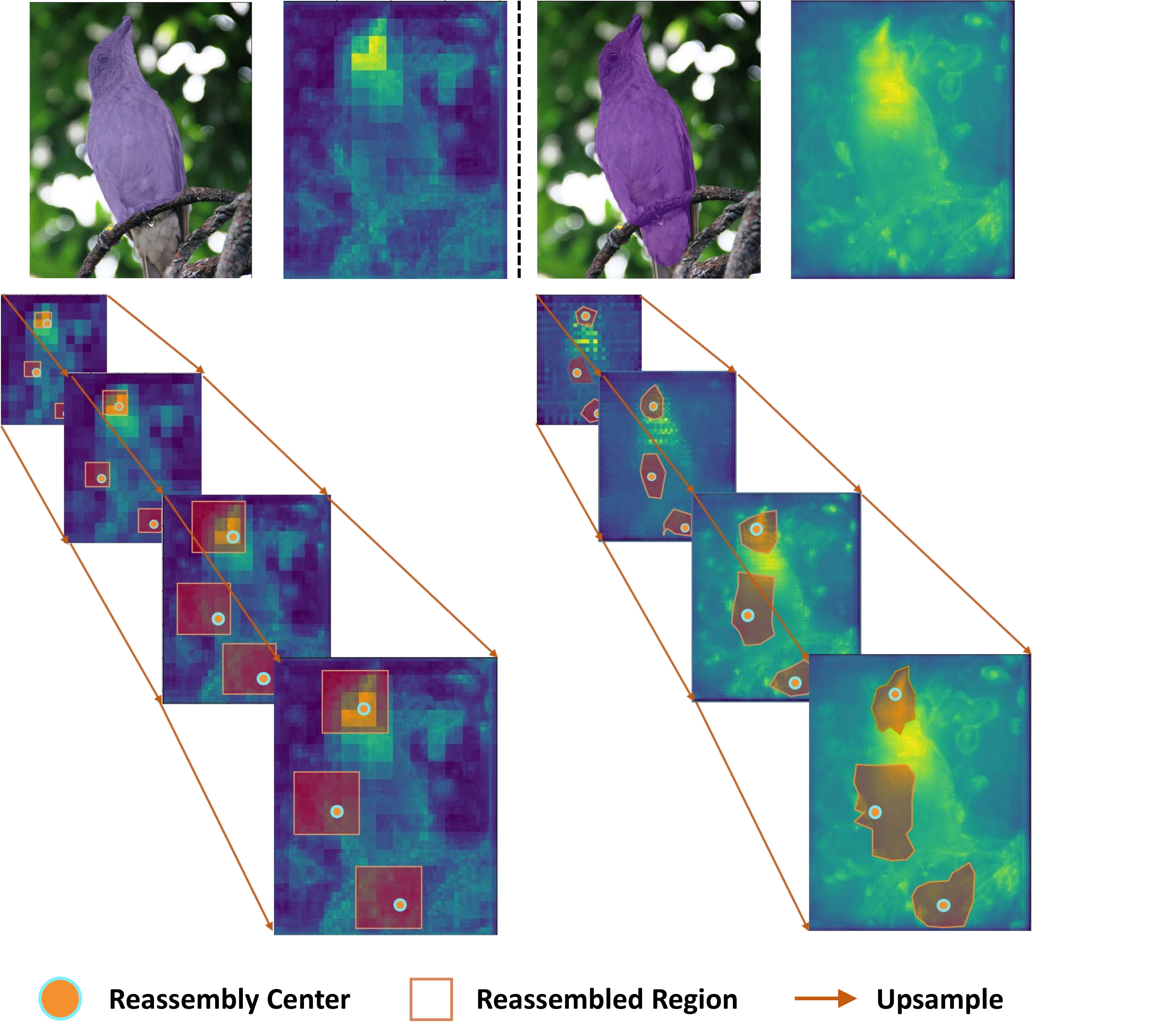}
	\caption{\small{\textbf{Illustration of \algname~working mechanism}. \textbf{Left:} Multi-level FPN features from Mask R-CNN (left to dotted line) and \textbf{Right:} Mask R-CNN with CARAFE (right to dotted line). 
			For sampled locations, this figure shows the accumulated reassembled regions in the top-down pathway of FPN.
			Information inside such a region is reassembled into the corresponding reassembly center.}
	}
	\label{fig:features}
	\vspace{-15pt}
\end{figure}

The most widely used feature upsampling operators are the nearest neighbor and bilinear interpolations, which adopt spatial distance between pixels to guide the upsampling process. 
However, nearest neighbor and bilinear interpolations only consider sub-pixel neighborhood, failing to capture the rich semantic information required by dense prediction tasks.
Another route toward adaptive upsampling is deconvolution~\cite{Noh_2015}. A deconvolution layer works as an inverse operator of a convolution layer, which learns a set of instance-agnostic upsampling kernels.
However, it has two major drawbacks.
Firstly, a deconvolution operator applies the same kernel across the entire image, regardless of the underlying content. This restricts its capability of responding to local variations. Second, it comes with a large number of parameters and thus heavy computational workload when a large kernel size is used. This makes it difficult to cover a larger region that goes beyond a small neighborhood, thus limiting its expressive power and performance.

In this work, we move beyond these limitations, and seek a feature upsampling operator that is capable of 1) aggregating information within large receptive field, 2) adapting to instance-specific contents on-the-fly, and 3) maintaining computation efficiency. 
To this end, we propose a lightweight yet highly effective operator, called \emph{Content-Aware ReAssembly of Features (CARAFE)}.
Specifically, \algname~ reassembles the features inside a predefined region centered at each location via a weighted combination, where the weights are generated in a content-aware manner.
Furthermore, there are multiple groups of such upsampling weights for each location. 
Feature upsampling is then accomplished by rearranging the generated features as a spatial block. 

Note that these spatially adaptive weights are not learned as network parameters. 
Instead, they are predicted on-the-fly, using a lightweight fully-convolutional module with softmax activation. 
Figure~\ref{fig:features} reveals the working mechanism of \algname.
After upsampled by \algname~, a feature map can represent the shape of an object more accurately, so that the model can predict better instance segmentation results. Our \algname~not only upsamples the feature map spatially, but also learns to enhance its discrimination.

To demonstrate the universal effectiveness of \algname~, we conduct comprehensive evaluations across a wide range of dense prediction tasks, \ie, object detection, instance segmentation, semantic segmentation, image inpainting, with mainstream architectures. 
\algname~can boost the performance of Faster RCNN~\cite{ren2015faster} by 1.2\% AP in object detection and Mask RCNN~\cite{he2017mask} by 1.3\% AP in instance segmentation on MS COCO~\cite{lin2014microsoft} test-dev 2018. 
\algname~further improves UperNet~\cite{xiao2018unified} by 1.8\% mIoU on ADE20k~\cite{zhou2017scene, zhou2018semantic} val in semantic segmentation, and improves Global\&Local~\cite{iizuka2017globally} by 1.1 dB of PSNR on Places~\cite{zhou2017places} val in image inpainting.
When upsampling an $H\times W$ feature map with 256 channels by a factor of two, the introduced computational overhead by \algname~is only $H*W*199k$ FLOPs, \vs, $H*W*1180k$ FLOPs of deconvolution.
The substantial gains on all the tasks demonstrate that \algname~is an effective and efficient feature upsampling operator that has great potential to serve as a strong building block for future research.


\section{Related Work}
\label{sec:related}

\noindent
\textbf{Upsampling Operators.} 
The most commonly used upsampling methods are nearest neighbor and bilinear interpolations.
These interpolations leverage distances to measure the correlations between pixels, and hand-crafted upsampling kernels are used in them.
In deep learning era, several methods are proposed to upsample a feature map using learnable operators.
For example, deconvolution~\cite{Noh_2015}, which is an inverse operator of a convolution, is the most famous among those learnable upsamplers. 
Pixel Shuffle~\cite{Shi_2016} proposes a different upsampler which reshapes depth on the channel space into width and height on the spatial space. 
Recently, \cite{mazzini2018guided} proposed guided upsampling (GUM), which performs interpolation by sampling pixels with learnable offsets. 
However, these methods either exploit contextual information in a small neighborhood, or require expensive computation to perform adaptive interpolation.
Within the realms of super-resolution and denoising, some other works~\cite{mildenhall2018burst,jo2018deep,hu2019meta} also explore using learnable kernels spatially in low-level vision.  
With a similar design spirit, here we demonstrate the effectiveness and working mechanism of content-aware feature reassembly for upsampling in several visual perception tasks, and provide a lightweight solution. 

\noindent
\textbf{Dense Prediction Tasks.}
Object detection is the task of localizing objects with bounding-boxes, instance segmentation further requires the prediction of instance-wise masks. 
Faster-RCNN~\cite{ren2015faster} introduces Region Proposal Network (RPN) for end-to-end training, which is further improved by the guided anchoring scheme~\cite{wang2019region}.
~\cite{lin2017feature,liu2018path,kong2018deep,zhao2019m2det,pang2019libra} exploits multi-scale feature
pyramids to deal with objects at different scales.
By adding extra mask prediction branches, Mask-RCNN~\cite{he2017mask} and
its variants~\cite{chen2019hybrid,huang2019mask} yield promising pixel-level results.
Semantic segmentation~\cite{liu2015semantic,li2017not} requires pixel-wise semantic prediction for given images. 
PSPNet~\cite{zhao2017pyramid} introduces spatial pooling at multiple grid scales.
and UperNet~\cite{xiao2018unified} designs a more generalized framework based on PSPNet.
Image or Video inpainting~\cite{yu2018generative,Xu_2019_CVPR,Xiong_2019_CVPR} is a classical problem to fill in the missing regions of the input pictures.
U-net~\cite{ronneberger2015u} is popular among recent works~\cite{iizuka2017globally,ulyanov2018deep}, and adopts multiple upsampling operators.
Liu \etal~\cite{liu2018image} introduce partial convolution layer to alleviate the influence of missing regions on the convolution layers.
Our \algname~demonstrates universal effectiveness across a wide range of dense prediction tasks.


\begin{figure*}[t]
	\centering
	\includegraphics[width=0.9\linewidth]{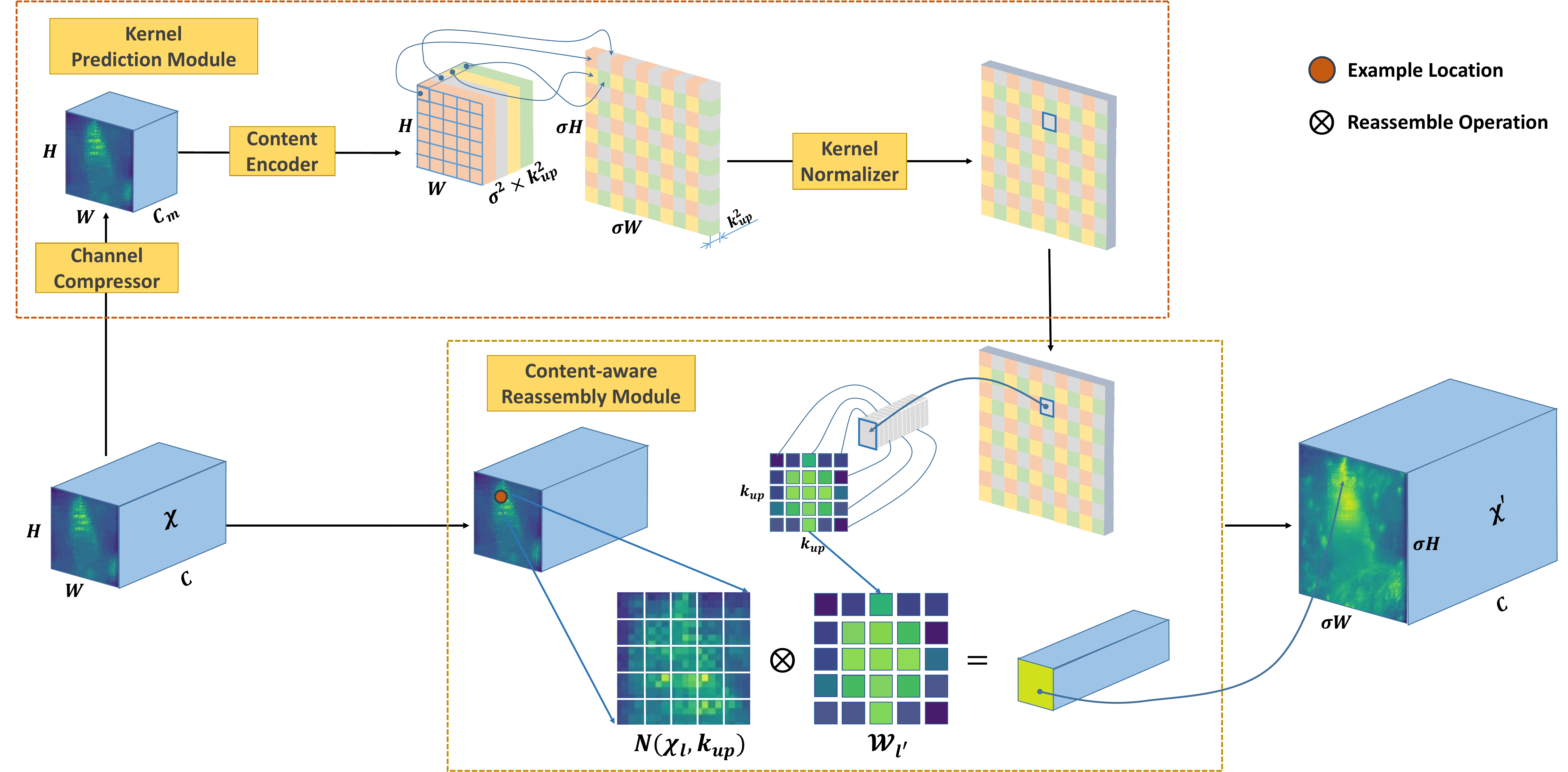}
	\caption{\small{\textbf{The overall framework of {\algname}}. {\algname} is composed of two key components, \ie, kernel prediction module and content-aware reassembly module. A feature map with size $C \times H\times W$ is upsampled by a factor of $\sigma(=2)$ in this figure.}}
	\label{fig:framework}
	\vspace{-15pt}
\end{figure*}

\section{Content-Aware ReAssembly of FEatures}
\label{sec:methodology}

Feature upsampling is a key operator in many modern convolutional network architectures developed for tasks including object detection, instance segmentation, and scene parsing.
In this work, we propose the content-aware reassembly of features (\algname) to upsample a feature map. On each location, \algname~can leverage the underlying content information 
to predict reassembly kernels and reassemble the features inside a predefined nearby region. Thanks to the content information, \algname~can use an adaptive and 
optimized reassembly kernel in different locations and achieve better performance than mainstream upsampling operators, \eg~interpolations or deconvolution.

\subsection{Formulation}
\label{subsec:formulation}
\algname~works as a reassembly operator with content-aware kernels.
It consists of two steps. The first step is to predict a reassembly kernel for
each target location according to its content, and the second step is to reassemble
the features with predicted kernels.
Given a feature map $\cX$ of size $C \times H\times W$ and an upsample ratio $\sigma$
(supposing $\sigma$ is an integer), \algname~will produce a new feature map
$\cX^\prime$ of size $C \times \sigma H\times\sigma W$.
For any target location $l^\prime=(i^\prime,j^\prime)$ of the output
$\cX^\prime$, there is a corresponding source location $l=(i,j)$ at the input $\cX$,
where $i=\left \lfloor i^\prime/\sigma \right \rfloor,j=\left \lfloor j^\prime/\sigma \right \rfloor$.
Here we denote $N(\cX_l, k)$ as the $k\times k$ sub-region of $\cX$ centered at the location $l$,
\ie, the neighbor of $\cX_l$.

In the first step, the \emph{kernel prediction module} $\psi$ predicts a
location-wise kernel $\cW_{l^\prime}$ for each location $l^\prime$,
based on the neighbor of $\cX_l$, as shown in Eqn.~\eqref{eq:kernel-prediction}.
The reassembly step is formulated as Eqn.~\eqref{eq:reassembly}, where $\phi$
is the \emph{content-aware reassembly module} that reassembles the neighbor of $\cX_l$
with the kernel $\cW_{l^\prime}$:
\begin{equation} \label{eq:kernel-prediction}
\cW_{l^\prime} = \psi(N(\cX_l, k_{encoder})).
\end{equation}
\begin{equation} \label{eq:reassembly}
\cX^\prime_{l^\prime} = \phi(N(\cX_l, k_{up}), \cW_{l^\prime}).
\end{equation}
We specify the details of $\psi$ and $\phi$ in the following parts.

\subsection{Kernel Prediction Module}
\label{subsec:kernel prediction module}

The kernel prediction module is responsible for generating the reassembly kernels in a content-aware manner.
Each source location on $\cX$ corresponds to $\sigma^2$ target locations on $\cX^\prime$. Each target locations requires a $k_{up}\times k_{up}$ reassembly kernel, where $k_{up}$ is the reassembly kernel size.
Therefore, this module will output the reassembly kernels of size $C_{up} \times H \times W$,
where $C_{up}=\sigma^2k_{up}^2$.

The kernel prediction module is composed of three submodules, \ie,
\emph{channel compressor}, \emph{content encoder} and \emph{kernel normalizer},
as shown in Figure~\ref{fig:framework}.
The channel compressor reduces the channel of the input feature map.
The content encoder then takes the compressed feature map as input and encodes
the content to generate reassembly kernels.
Lastly, the kernel normalizer applies a softmax function to each reassembly kernel.
The three submodules are explained in detail as follows.

\noindent\textbf{Channel Compressor}.
We adopt a $1\times 1$ convolution layer to compress the input feature channel from $C$ to $C_m$.
Reducing the channel of input feature map leads to less parameters and
computational cost in the following steps, making \algname~more efficient.
It is also possible to use larger kernel sizes for the content encoder under the same budget.
Experimental results show that reducing the feature channel in an acceptable
range will not harm the performance.

\noindent\textbf{Content Encoder}.
We use a convolution layer of kernel size $k_{encoder}$ to generate reassembly
kernels based on the content of input features.
The parameters of the encoder is $k_{encoder}\times k_{encoder}\times C_m\times C_{up}$.
Intuitively, increasing $k_{encoder}$ can enlarge the receptive field of the
encoder, and exploits the contextual information within a larger region,
which is important for predicting the reassembly kernels.
However, the computational complexity grows with the square of the kernel size,
while the benefits from a larger kernel size do not.
An empirical formula $k_{encoder} = k_{up}-2$ is a good trade-off between
performance and efficiency through our study in Section~\ref{subsec:ablation}.

\noindent\textbf{Kernel Normalizer}.
Before being applied to the input feature map, each $k_{up}\times k_{up}$
reassembly kernel is normalized with a softmax function spatially.
The normalization step forces the sum of kernel values to 1, which is a soft selection across a local region.
Due to the kernel normalizer, \algname~does not perform any rescaling and change
the mean values of the feature map, that is why our proposed operator is named
the reassembly of features.

\subsection{Content-aware Reassembly Module}
\label{subsec:content-aware reassembly module}

With each reassembly kernel $\cW_{l^\prime}$, the content-aware reassembly module
will reassemble the features within a local region via the function $\phi$.
We adopt a simple form of $\phi$ which is just a weighted sum operator.
For a target location $l^\prime$ and the corresponding square region $N(\cX_l, k_{up})$ centered at $l=(i,j)$, the reassembly
is shown in Eqn.~\eqref{eq:equal_3}, where $r = \left \lfloor k_{up} / 2 \right \rfloor$:
\begin{equation} \label{eq:equal_3}
\begin{aligned}
\cX^\prime_{l^\prime}= 
\sum_{n = -r}^{r}\sum_{m = -r}^{r} \cW_{l^\prime(n,m)} \cdot \cX_{(i+n, j+m)}.
\end{aligned}
\end{equation}

With the reassembly kernel, each pixel in the region of $N(\cX_l, k_{up})$
contributes to the upsampled pixel $l^\prime$ differently, based on the content
of features instead of distance of locations.
The semantics of the reassembled feature map can be stronger than the original
one, since the information from relevant points in a local region can be more attended.

\subsection{Relation to Previous Operators}
Here we discuss the relations between \algname~and dynamic filter~\cite{brab2016dynamic}, spatial attention~\cite{chen2017sca}, spatial transformer~\cite{jaderberg2015spatial} and deformable convolution~\cite{Dai_2017}, which share similar design philosophy but with different focuses.

\noindent
\textbf{Dynamic Filter.}
Dynamic filter generates instance-specific convolutional filters conditioned on the input of the network, and then applies the predicted filter on the input.
Both dynamic filter and \algname~are content-aware operators, but a fundamental difference between them lies at their kernel generation process. 
Specifically, dynamic filter works as a two-step convolution, where the additional filter prediction layer and filtering layer require heavy computation.
On the contrary, \algname~is simply a reassembly of features in local regions, without learning the feature transformation across channels.
Supposing the channels of input feature map is $C$ and kernel size of the
filter is $K$, then the predicted kernel parameters for each location is
$C\times C\times K\times K$ in dynamic filter. For \algname, the kernel parameter is only $K\times K$. Thus, it is more efficient in memory and speed.

\noindent
\textbf{Spatial Attention.}
Spatial attention predicts an attention map with the same size as the input feature, and then rescales the feature map on each location.
Our \algname~reassembles the features in a local region by weighted sum.
In summary, spatial attention is a rescaling operator with point-wise guidance while \algname~is a reassembly operator with region-wise local guidance.
Spatial attention can be seen as a special case of \algname~where the reassembly kernel size is 1, regardless of the kernel normalizer.

\noindent
\textbf{Spatial Transformer Networks (STN).}
STN predicts a global parametric transformation conditioned on the input feature map and warps the feature via the transformation.
However, this global parametric transformation assumption is too strong to represent complex spatial variance; and STN is known to be hard to train.
Here, \algname~uses the location-specific reassembly to handle the spatial relations, which enables more flexible local geometry modeling.

\noindent
\textbf{Deformable Convolutional Networks (DCN).}
DCN also adopts the idea of learning geometric transformation and combines it with the regular convolution layers. 
It predicts kernel offsets other than using grid convolution kernels.
Similar to dynamic filter, it is also a heavy parametric operator with 24 times more computational cost than \algname.
It is also known to be sensitive to parameter initialization.


\section{Applications of \algname}
\label{sec:applications}

\algname~can be seamlessly integrated into existing frameworks where
upsampling operators are needed.
Here we present some applications in mainstream dense prediction tasks. With negligible additional parameters, \algname~benefits state-of-the-art methods in
both high-level and low-level tasks, such as object detection, instance segmentation, semantic segmentation and image inpainting.

\subsection{Object Detection and Instance Segmentation}
\label{subsec:application in object detection and instance segmentation}

\begin{figure}
	\centering
	\includegraphics[width=\linewidth]{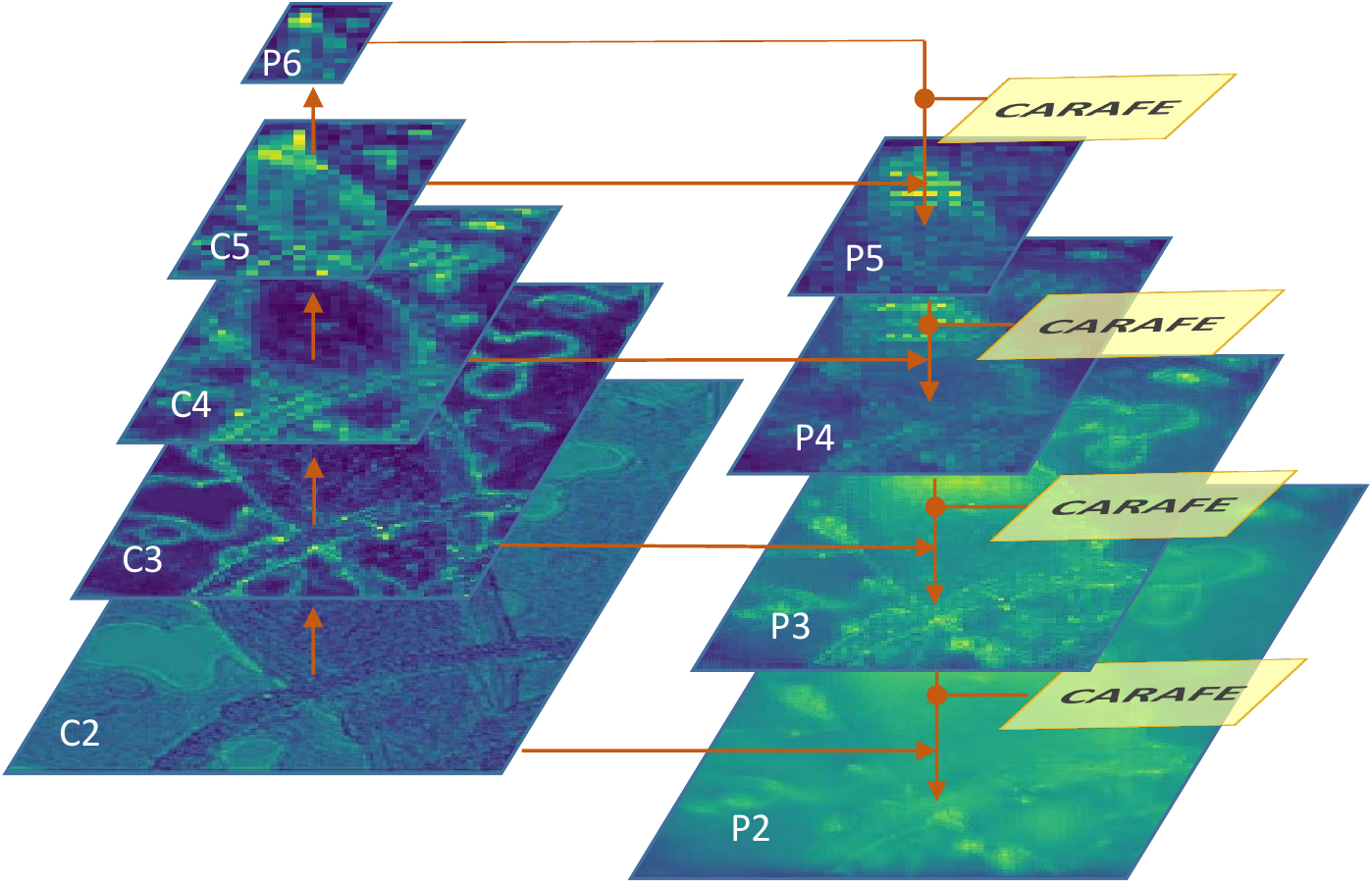}
	\caption{\small{\textbf{FPN architecture with \algname}. \algname~upsamples a feature map by a factor of 2 in the top-down pathway. It is integrated into FPN by seamlessly substituting the nearest neighbor interpolation.}}
	\label{fig:fpn}
	\vspace{-10pt}
\end{figure}

Feature Pyramid Network (FPN) is an important and effective architecture in the
field of object detection and instance segmentation. It significantly improves
the performance of popular frameworks like Faster R-CNN and Mask R-CNN.
FPN constructs feature pyramids of strong semantics with the top-down pathway
and lateral connections. In the top-down pathway, a low-resolution feature map
is firstly upsampled by 2x with the nearest neighbor interpolation and then fused with a
high-resolution one, as shown in Figure~\ref{fig:fpn}.

We propose to substitute the nearest neighbor interpolation in all the feature levels
with \algname. This modification is smooth and no extra change is required.
In addition to the FPN structure, Mask R-CNN adopts a deconvolution layer at
the end of mask head. It is used to upsample the predicted digits from
$14\times14$ to $28\times28$, to obtain finer mask predictions.
We can also use \algname~to replace the deconvolution layer, resulting in
even less computational cost.

\subsection{Semantic Segmentation}
\label{subsec:application in semantic segmentation}

Semantic segmentation requires the model to output per-pixel level predictions
on the whole image, so that high-resolution feature maps are usually preferred.
Upsampling is widely adopted to enlarge feature maps and fuse the semantic
information of different levels in this task.
UperNet is a strong baseline for semantic segmentation. It uses upsampling in
the following three components, \ie, PPM, FPN, FUSE. We adopt \algname~instead of their original upsamplers.

\noindent\textbf{Pyramid Pooling Module (PPM).}
PPM is the key component in PSPNet that hierarchically down-samples an input
feature map into multiple scales $\{1\times1, 2\times2, 3\times3, 6\times6\}$,
and then upsamples them back to the original sizes with bilinear interpolation.
The features are finally fused with the original feature by concatenation.
Since the upsampling ratio is very large, we adopt a two-step strategy with
\algname~as a trade-off between performance and efficiency.
Firstly we upsamples the $\{1\times1, 2\times2, 3\times3, 6\times6\}$
features to half the size of the original feature map with bilinear interpolation,
and then use \algname~to further upsample them by 2x.

\noindent\textbf{Feature Pyramid Network (FPN).}
Similar to detection models, UperNet also adopts FPN to enrich the feature
semantics. It only has four different feature levels \{P2, P3, P4, P5\} with strides \{4, 8, 16, 32\}.
We replace the upsampling operators in the same way as Section~\ref{subsec:application in object detection and instance segmentation}.

\noindent\textbf{Multi-level Feature Fusion (FUSE).}
UperNet proposes a multi-level feature fusion module after the FPN.
It upsamples P3, P4, P5 to the same size as P2 by bilinear interpolation
and then fuses these features from different levels by concatenation.
The process is equivalent to a sequential upsampling-concatenation that first
upsamples P5 to P4 and concatenates them, and then upsamples the concatenated
feature map to P3 and so on. We replace the sequential bilinear upsampling here with \algname.

\subsection{Image Inpainting}
\label{subsec:application in image inpainting}
The U-net architecture is popular among recent proposed image inpainting
methods, such as Global\&Local~\cite{iizuka2017globally} and Partial Conv~\cite{liu2018image}.
There are two upsampling operators in the second half of the network.
We simply replace the two upsampling layers with \algname~and evaluate the 
performance.
As for Partial Conv, we can conveniently keep the mask propagation in
\algname~by updating the mask with our content-aware reassembly kernels.


\begin{table*}[t]
	\centering
	\caption{\small{Detection and Instance Segmentation results on MS COCO 2018 \emph{test-dev}.}}
	\addtolength{\tabcolsep}{-2pt}
	\small{
		\begin{tabular}{lllllllll}
			\hline
			\multirow{2}{*}{Method} & \multirow{2}{*}{Backbone} & \multirow{2}{*}{Task} & \multirow{2}{*}{AP} & \multirow{2}{*}{$\text{AP}_{50}$} & \multirow{2}{*}{$\text{AP}_{75}$} & \multirow{2}{*}{$\text{AP}_{S}$} & \multirow{2}{*}{$\text{AP}_{M}$} & \multirow{2}{*}{$\text{AP}_{L}$} \\
			&  &  &  &  &  &  &  &  \\ \hline
			Faster R-CNN & ResNet-50 & BBox & 36.9 & 59.1 & 39.7 & 21.5 & 40.0 & 45.6 \\
			Faster R-CNN w/ \algname & ResNet-50 & BBox & \textbf{38.1} & \textbf{60.7} & \textbf{41.0} & \textbf{22.8} & \textbf{41.2} & \textbf{46.9} \\ \hline
			\multirow{2}{*}{Mask R-CNN} & ResNet-50 & BBox & 37.8 & 59.7 & 40.8 & 22.2 & 40.7 & 46.8 \\
			& ResNet-50 & Segm & 34.6 & 56.5 & 36.8 & 18.7 & 37.3 & 45.1 \\
			\multirow{2}{*}{Mask R-CNN w/ \algname} & ResNet-50 & BBox & \textbf{38.8} & \textbf{61.2} & \textbf{42.1} & \textbf{23.2} & \textbf{41.7} & \textbf{47.9} \\
			& ResNet-50 & Segm & \textbf{35.9} & \textbf{58.1} & \textbf{38.2} & \textbf{19.8} & \textbf{38.6} & \textbf{46.5} \\ \hline
		\end{tabular}
	}
	\label{tab:det-results}
	\vspace{-10pt}
\end{table*}

\section{Experiments}
\label{sec:experiments}

\subsection{Experimental Settings}

\noindent
\textbf{Datasets \& Evaluation Metrics.}
We evaluate \algname~on several important dense prediction benchmarks.
We use the \emph{train} split for training and evaluate the performance on
the \emph{val} split for all these datasets by default.

\noindent\emph{Object Detection and Instance Segmentation}.
We perform experiments on the challenging MS COCO 2017 dataset.
Results are evaluated with the standard COCO metric, \ie mAP of IoUs from 0.5 to 0.95.

\noindent\emph{Semantic Segmentation}.
We adopt the ADE20k benchmark to evaluate our method in the semantic segmentation task.
Results are measured with mean IoU (mIoU) and Pixel Accuracy (P.A.),
which respectively indicates the average IoU between predictions and ground
truth masks and per-pixel classification accuracy.

\noindent\emph{Image Inpainting}.
Places dataset is adopted for image inpainting.
We use L1 error (lower is better) and PSNR (higher is better) as evaluation metrics.

\noindent
\textbf{Implementation Details.}
If not otherwise specified, \algname~adopts a fixed set of hyper-parameters in
experiments, where $C_m$ is 64 for the channel compressor and
$k_{encoder}=3$, $k_{up}=5$ for the content encoder. See more implementation details in supplementary materials.

\noindent\emph{Object Detection and Instance Segmentation}.
We evaluate \algname~on Faster RCNN and Mask RCNN with the ResNet-50 w/ FPN backbone, 
and follow the 1x training schedule settings as Detectron~\cite{Detectron2018} and MMDetection~\cite{mmdetection}.

\noindent\emph{Semantic Segmentation}.
We use the official implementation of UperNet\footnote{\href{https://github.com/CSAILVision/semantic-segmentation-pytorch}{https://github.com/CSAILVision/semantic-segmentation-pytorch}} and adopt the same experiment settings.

\noindent\emph{Image Inpainting}
We adopt Global\&Local~\cite{iizuka2017globally} and Paritial Conv~\cite{liu2018image} as baseline methods to evaluate \algname.

\subsection{Benchmarking Results}
\label{subsec:results}

\noindent
\textbf{Object Detection \& Instance Segmentation.}
We first evaluate our method by substituting the nearest neighbor interpolation in FPN with
\algname~for both Faster RCNN and Mask RCNN, and the deconvolution layer in the
mask head for Mask RCNN.
As shown in Table~\ref{tab:det-results}, \algname~improves Faster RCNN by 1.2\%
on bbox AP, and Mask RCNN by 1.3\% on mask AP.
The improvements of $\text{AP}_{S}$, $\text{AP}_{M}$, $\text{AP}_{L}$ are all above 1\% AP, which suggests that it is beneficial for various object scales.

\begin{figure*}[t]
	\centering
	\includegraphics[width=\linewidth]{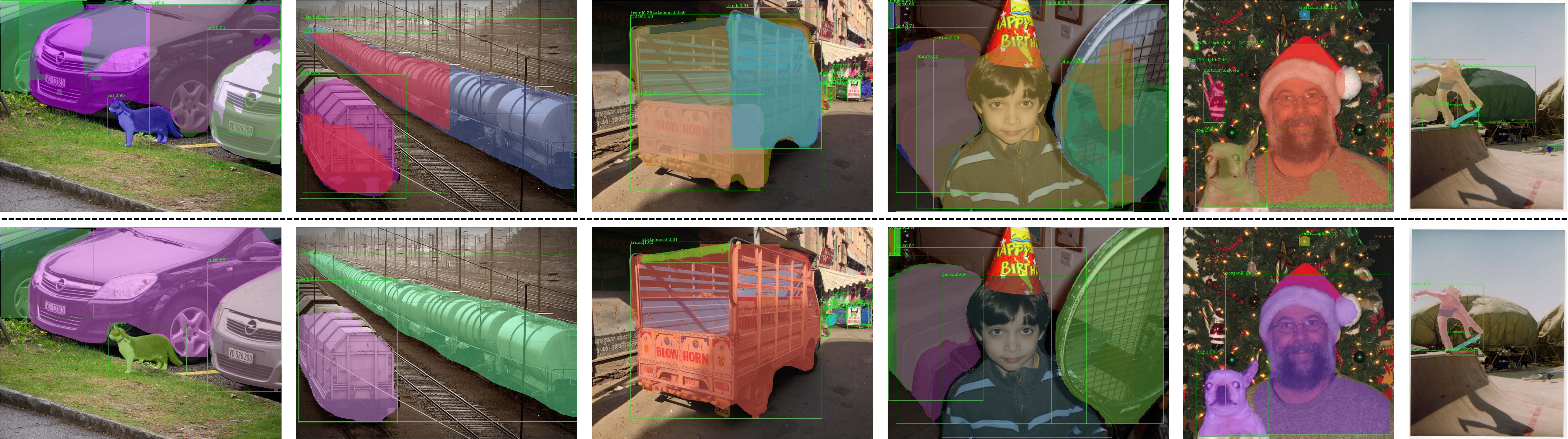}
	\caption{\small{Comparison of instance segmentation results between baseline (top row) and \algname~ (bottom row) on COCO 2017 val.}}
	\label{fig:instance_results}
	\vspace{-15pt}
\end{figure*}

Our encouraging performance is supported by the qualitative results as shown
in Figure~\ref{fig:features}.
We visualize the feature maps in the top-down pathway of FPN and compare \algname~
with the baseline, \ie, nearest neighbor interpolation.
It is obvious that with the content-aware reassembly, the feature maps are more
discriminative and a more accurate mask for the object is predicted.
In Figure~\ref{fig:instance_results}, we show some examples of instance
segmentation results comparing the baseline and \algname.

\begin{table}[t]
	\centering
	\caption{\small{Detection results with Faster RCNN. Various upsampling methods are used in FPN. N.C., B.C., P.S. and S.A. 
			indicate Nearest + Conv, Bilinear + Conv, Pixel Shuffle and Spatial Attention, respectively.}}
	\addtolength{\tabcolsep}{-4pt}
	\small{
		\begin{tabular}{lllllllll}
			\hline
			\multirow{2}{*}{Method} & \multirow{2}{*}{AP} & \multirow{2}{*}{$\text{AP}_{50}$} & \multirow{2}{*}{$\text{AP}_{75}$} & \multirow{2}{*}{$\text{AP}_{S}$} & \multirow{2}{*}{$\text{AP}_{M}$} & \multirow{2}{*}{$\text{AP}_{L}$} & \multirow{2}{*}{FLOPs} & \multirow{2}{*}{Params} \\
			&  &  &  &  &  &  &  & \\ \hline
			Nearest & 36.5 & 58.4 & 39.3 & 21.3 & 40.3 & 47.2 & 0 & 0\\
			Bilinear & 36.7 & 58.7 & 39.7 & 21.0 & 40.5 & 47.5 & 8k & 0\\
			N.C. & 36.6 & 58.6 & 39.5 & 21.4 & 40.3 & 46.4 & 4.7M & 590k\\
			B.C. & 36.6 & 58.7 & 39.4 & 21.6 & 40.6 & 46.8 & 4.7M & 590k\\
			Deconv~\cite{Noh_2015} & 36.4 & 58.2 & 39.2 & 21.3 & 39.9 & 46.5 & 1.2M & 590k\\
			P.S.\cite{Shi_2016} & 36.5 & 58.8 & 39.1 & 20.9 & 40.4 & 46.7 & 4.7M & 2.4M\\
			GUM\cite{mazzini2018guided} & 36.9 & 58.9 & 39.7 & 21.5 & 40.6 & 48.1 & 1.1M & 132k\\
			S.A.\cite{chen2017sca} & 36.9 & 58.8 & 39.8 & 21.7 & 40.8 & 47.0 & 28k & 2.3k\\
			\algname & \textbf{37.8} & \textbf{60.1} & \textbf{40.8} & \textbf{23.1} & \textbf{41.7} & \textbf{48.5} & 199k & 74k\\ \hline
		\end{tabular}
	}
	\label{tab:upsample compare}
	\vspace{-15pt}
	
\end{table}

To investigate the effectiveness of different upsampling operators, we perform
extensive experiments on Faster RCNN by using different operators to perform upsampling in FPN.
Results are illustrated in Table~\ref{tab:upsample compare}.
For `N.C.' and `B.C.', which respectively indicate `Nearest + Conv' and `Bilinear + Conv', we add an extra $3\times3$
convolution layer after the corresponding upsampling.
`Deconv', `Pixel Shuffle' (indicated as `P.S.'), `GUM' are three representative learning based upsampling methods.
We also compare `Spatial Attention' here, indicated as `S.A.'.
\algname~achieves the best AP among all these upsampling operators, the
FLOPs and parameters are relatively small, which proves it is both effective and efficient.
The results of `Nearest + Conv' and `Bilinear + Conv' show that extra parameters
do not lead to a significant gain.
`Deconv', `Pixel Shuffle', `GUM' and `Spatial Attention' obtain inferior performance to \algname,
indicating that the design of effective upsampling operators is critical.

\begin{table}[t]
	\centering
	\caption{\small{Instance Segmentation results with Mask RCNN. Various upsampling methods are used in mask head.}}
	\addtolength{\tabcolsep}{-2pt}
	\small{
		\begin{tabular}{lllllll}
			\hline
			\multirow{2}{*}{Method} & \multirow{2}{*}{AP} & \multirow{2}{*}{$\text{AP}_{50}$} & \multirow{2}{*}{$\text{AP}_{75}$} & \multirow{2}{*}{$\text{AP}_{S}$} & \multirow{2}{*}{$\text{AP}_{M}$} & \multirow{2}{*}{$\text{AP}_{L}$} \\
			&  &  &  &  &  &  \\ \hline
			Nearest & 32.7 & 55.0 & 34.8 & 17.7 & 35.9 & 44.4 \\
			Bilinear & 34.2 & 55.9 & 36.4 & 18.5 & 37.5 & 46.2 \\
			Deconv & 34.2 & 55.5 & 36.3 & 17.6 & 37.8 & 46.7 \\
			Pixel Shuffle & 34.4 & 56.0 & 36.6 & \textbf{18.5} & 37.6 & 47.5 \\
			GUM & 34.3 & 55.7 & 36.5 & 17.6 & 37.6 & 46.9 \\
			S.A. & 34.1 & 55.6 & 36.5 & 17.6 & 37.4 & 46.6 \\
			\algname & \textbf{34.7} & \textbf{56.2} & \textbf{37.1} & 18.2 & \textbf{37.9} & \textbf{47.5} \\ \hline
		\end{tabular}
	}
	\label{tab:mask head compare}
	\vspace{-7pt}
	
\end{table}

Besides FPN which is a pyramid feature fusion structure, we also explore different
upsampling operators in the mask head.
In typical Mask R-CNN, a deconvolution layer is adopted to upsample the RoI
features by 2x.
For a fair comparison, we do not make any changes to FPN, and only replace the
deconvolution layer with various operators.
Since we only modify the mask prediction branch, performance is reported in
terms of mask AP, as shown in Table~\ref{tab:mask head compare}.
\algname~achieves the best performance in instance segmentation among these methods.

\begin{table}[t]
	\centering
	\caption{\small{Detection and Instance Segmentation results with Mask RCNN via adopting \algname~in FPN and mask head respectively.
			M.H. indicates using \algname~in mask head.}}
	\addtolength{\tabcolsep}{-2pt}
	\small{
		\begin{tabular}{lllllllll}
			\hline
			\multirow{2}{*}{FPN} & \multirow{2}{*}{M.H.} & \multirow{2}{*}{Task} & \multirow{2}{*}{AP} & \multirow{2}{*}{$\text{AP}_{50}$} & \multirow{2}{*}{$\text{AP}_{75}$} & \multirow{2}{*}{$\text{AP}_{S}$} & \multirow{2}{*}{$\text{AP}_{M}$} & \multirow{2}{*}{$\text{AP}_{L}$} \\
			&  &  &  &  &  &  &  &  \\ \hline
			\multirow{2}{*}{} & \multirow{2}{*}{} & Bbox & 37.4 & 59.1 & 40.3 & 21.2 & 41.2 & 48.5 \\
			&  & Segm & 34.2 & 55.5 & 36.3 & 17.6 & 37.8 & 46.7 \\ \hline
			\multirow{2}{*}{\checkmark} & \multirow{2}{*}{} & Bbox & 38.6 & 60.7 & \bf{42.2} & 23.2 & 42.1 & 49.5 \\
			&  & Segm & 35.2 & 57.2 & 37.5 & 19.3 & 38.3 & 47.6 \\ \hline
			\multirow{2}{*}{} & \multirow{2}{*}{\checkmark} & Bbox & 37.3 & 59.0 & 40.2 & 21.8 & 40.8 & 48.6 \\
			&  & Segm & 34.7 & 56.2 & 37.1 & 18.2 & 37.9 & 47.5 \\ \hline
			\multirow{2}{*}{\checkmark} & \multirow{2}{*}{\checkmark} & Bbox & \bf{38.6} & \bf{60.9} & 41.9 & \bf{23.4} & \bf{42.3} & \bf{49.8} \\
			&  & Segm & \bf{35.7} & \bf{57.6} & \bf{38.1} & \bf{19.4} & \bf{39.0} & \bf{48.7} \\ \hline
		\end{tabular}
		\vspace{-15pt}
	}
	\label{tab:mask rcnn}
	
\end{table}

In Table~\ref{tab:mask rcnn}, we report the object detection and instance segmentation
results of adopting \algname~in FPN and mask head on Mask RCNN respectively.
Consistent improvements are achieved in these experiments.

\noindent
\textbf{Semantic Segmentation.}
We replace the upsamplers in UperNet with \algname~and evaluate the results on ADE20k benchmark.
As shown in Table~\ref{tab:semantic segmentation}, \algname~improves the mIoU
by a large margin from 40.44\% to 42.23\% with single scale testing.
Note that UperNet with \algname~also achieves better performance than recent strong baselines such as PSPNet\cite{zhao2017pyramid} and PSANet\cite{zhao2018psanet}.

\begin{table}[t]
	\centering
	\caption{\small{Semantic Segmentation results on ADE20k val. Single scale testing is 
			used in our experiments. P.A. indicates Pixel Accuracy.}}
	\addtolength{\tabcolsep}{-2pt}
	\small{
		\begin{tabular}{llll}
			\hline
			\multirow{2}{*}{Method} & \multirow{2}{*}{Backbone} & \multirow{2}{*}{mIoU} & \multirow{2}{*}{P.A.} \\
			&  &  &  \\ \hline
			PSPNet & ResNet-50 & 41.68 & 80.04 \\
			PSANet & ResNet-50 & 41.92 & 80.17 \\
			UperNet\tablefootnote{We report the performance in model zoo of the official implementation.} & ResNet-50 & 40.44 & 79.80 \\
			UperNet w/ CARAFE & ResNet-50 & \textbf{42.23} & \textbf{80.34} \\ \hline
		\end{tabular}
	}
	\label{tab:semantic segmentation}
	\vspace{-7pt}
	
\end{table}

We perform a step-by-step study to inspect the effectiveness of modifying different
components in UperNet, as described in Section~\ref{subsec:application in semantic segmentation}.
Results in Table~\ref{tab:semantic segmentation by step} show that \algname~is
helpful for all the three components and the combination of them results in
further gains.

\begin{table}[t]
	\centering
	\caption{\small{Effects of adopting CARAFE in each component of UperNet.}}
	\vspace{-10pt}
	\addtolength{\tabcolsep}{-2pt}
	\small{
		\begin{tabular}{lllll}
			\hline
			\multirow{2}{*}{PPM} & \multirow{2}{*}{FPN} & \multirow{2}{*}{FUSE} & \multirow{2}{*}{mIoU} & \multirow{2}{*}{P.A.} \\
			&  &  &  &  \\ \hline
			\checkmark &  &  & 40.85 & 79.97 \\
			& \checkmark &  & 40.79 & 80.01 \\
			&  & \checkmark & 41.06 & 80.23 \\
			\checkmark & \checkmark &  & 41.55 & 80.30 \\
			\checkmark &  & \checkmark & 42.01 & 80.11 \\
			& \checkmark & \checkmark & 41.93 & \bf{80.34} \\
			\checkmark & \checkmark & \checkmark & \bf{42.23} & \bfseries{80.34} \\ \hline
		\end{tabular}
	}
	\label{tab:semantic segmentation by step}
	\vspace{-15pt}
\end{table}

\noindent
\textbf{Image Inpainting.}
We show that \algname~is also effective in low-level tasks such as image inpainting.
By replacing the upsampling operators with \algname~in two strong baselines
Global\&Local~\cite{iizuka2017globally} and Partial Conv~\cite{liu2018image}, we observe significant
improvements for both methods.
As shown in Table~\ref{tab:inpainting}, our method improves two baselines by
1.1 dB and 0.2 dB on the PSNR metric.

\begin{table}[t]
	\small
	\centering
	\caption{\small{Image inpainting results on Places val.}}
	\vspace{-4pt}
	\addtolength{\tabcolsep}{-2pt}
	\small{
		\begin{tabular}{lll}
			\hline
			\multirow{2}{*}{Method}& \multirow{2}{*}{L1(\%)} & \multirow{2}{*}{PSNR(dB)} \\
			&  &  \\ \hline
			Global\&Local & 6.78 & 19.58 \\
			Partial Conv & 5.96 & 20.78 \\
			Global\&Local w/ \algname & 6.00 & 20.71 \\
			Partial Conv w/ \algname & \bf{5.72} & \bf{20.98} \\ \hline
		\end{tabular}
	}
	\label{tab:inpainting}
	\vspace{-15pt}
\end{table}

\begin{figure*}
	\centering
	\includegraphics[width=\linewidth]{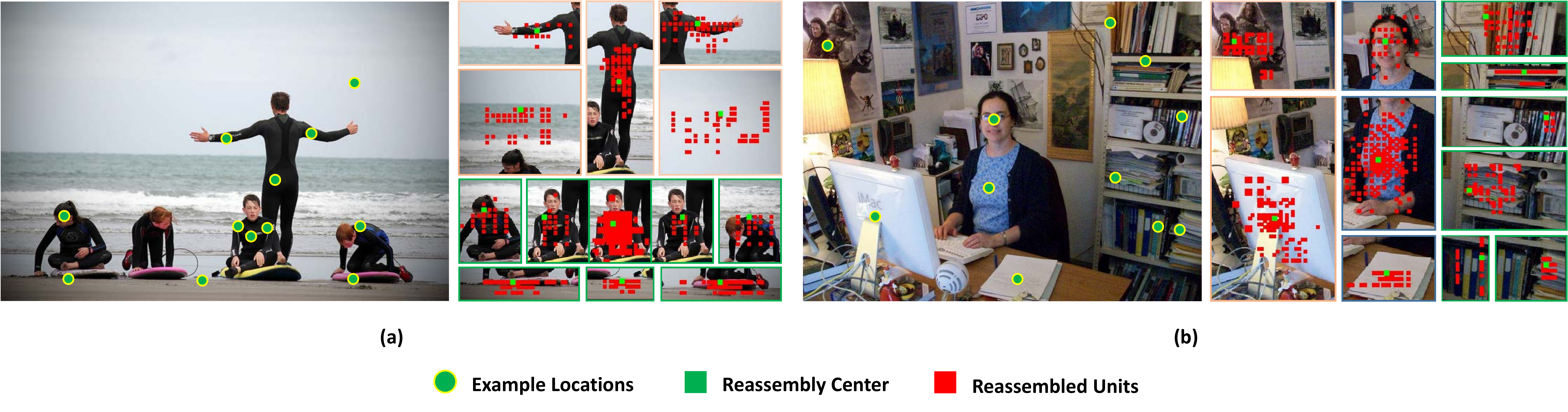}
	\caption{\small{\algname~performs content-aware reassembly when upsampling a feature map. 
			Red units are reassembled into the green center unit by \algname~in the top-down pathway of a FPN structure. }}
	\vspace{-15pt}
	\label{fig:arrow}
\end{figure*}

\subsection{Ablation Study \& Further Analysis}
\label{subsec:ablation}

\noindent
\textbf{Model Design \& Hyper-parameters.}
We investigate the influence of hyper-parameters in the model design, \ie,
the compressed channels $C_m$, encoder kernel size $k_{encoder}$ and reassembly
kernel size $k_{up}$. We also test different normalization methods in the
kernel normalizer.
We perform the ablation study of the designs and settings on Faster RCNN with
a ResNet-50 backbone, and evaluate the results on COCO 2017 val.

Towards an efficient design, we first analyze the computational complexity measured by FLOPs.
When upsampling the feature map with input channel $C_{in}$ by a factor of $\sigma$,
the per pixel FLOPs of \algname~is computed as
$2(C_{in}+1)C_{m}+2(C_{m}k_{encoder}^2+1)\sigma^2k_{up}^2+2\sigma^2k_{up}^2C_{in}$,
referring to~\cite{molchanov2016pruning}.

We experiment with different values of $C_m$ in the channel compressor.
In addition, we also try removing the channel compressor module, which means
the content encoder directly uses input features to predict reassembly kernels.
Experimental results in Table~\ref{tab:mid channels} show that compress $C_m$
down to 64 leads to no performance decline, while being more efficient.
A further smaller $C_m$ will result in a slightly drop of the performance.
With no channel compressor, it can achieve the same performance, which proves that
the channel compressor can speed up the kernel prediction without harming the performance.
Based on the above results, we set $C_m$ to 64 by default as a trade-off between
performance and efficiency.

\begin{table}[t]
	\centering
	\caption{\small{Ablation study of various compressed channels $C_m$. N/A means \emph{channel compressor} is removed. }}
	\addtolength{\tabcolsep}{-2pt}
	\small{
		\begin{tabular}{lllllll}
			\hline
			\multirow{2}{*}{$C_m$} & \multirow{2}{*}{AP} & \multirow{2}{*}{$\text{AP}_{50}$} & \multirow{2}{*}{$\text{AP}_{75}$} & \multirow{2}{*}{$\text{AP}_{S}$} & \multirow{2}{*}{$\text{AP}_{M}$} & \multirow{2}{*}{$\text{AP}_{L}$} \\
			&  &  &  &  &  &  \\ \hline
			16 & 37.6 & 60.1 & 40.6 & 22.7 & 41.6 & 48.4 \\
			32 & 37.7 & 60.3 & 40.7 & 22.8 & 41.2 & 49.0 \\
			64 & \bf{37.8} & 60.1 & 40.8 & 23.1 & 41.7 & 48.5 \\
			128 & \bf{37.8} & 60.1 & 40.8 & 22.4 & 41.7 & 48.7 \\
			256 & \bf{37.8} & 60.4 & 40.8 & 22.7 & 41.3 & 48.8 \\
			N/A & \bf{37.8} & 60.3 & 40.8 & 22.9 & 41.5 & 48.7 \\ \hline
		\end{tabular}
		\vspace{-10pt}
	}
	\label{tab:mid channels}
	
\end{table}

We then investigate the influence of $k_{encoder}$ and $k_{up}$.
Intuitively, increasing $k_{up}$ also requires a larger $k_{encoder}$,
since the content encoder needs a large receptive field to predict a large
reassembly kernel.
As illustrated in Table~\ref{tab:k study}, increasing $k_{encoder}$ and $k_{up}$
at the same time can boost the performance, while just enlarging one of them will not.
We summarize an empirical formula that $k_{encoder}=k_{up}-2$, which is a good
choice in all the settings.
Though adopting larger kernel size is shown helpful, we set $k_{up}=5$ and
$k_{encoder}=3$ by default as a trade-off between performance and efficiency.

\begin{table}[t]
	\centering
	\caption{\small{Detection results with various encoder kernel size $k_{encoder}$ and reassembly
			kernel size $k_{up}$.} }
	\addtolength{\tabcolsep}{-2pt}
	\small{
		\begin{tabular}{llllllll}
			\hline
			\multirow{2}{*}{$k_{encoder}$} & \multirow{2}{*}{$k_{up}$} & \multirow{2}{*}{AP} & \multirow{2}{*}{$\text{AP}_{50}$} & \multirow{2}{*}{$\text{AP}_{75}$} & \multirow{2}{*}{$\text{AP}_{S}$} & \multirow{2}{*}{$\text{AP}_{M}$} & \multirow{2}{*}{$\text{AP}_{L}$} \\
			&  &  &  &  &  &  &  \\ \hline
			1 & 3 & 37.3 & 59.6 & 40.5 & 22.0 & 40.7 & 48.1 \\
			1 & 5 & 37.3 & 59.9 & 40.0 & 22.3 & 41.1 & 47.3 \\
			3 & 3 & 37.3 & 59.7 & 40.4 & 22.1 & 40.8 & 48.3 \\
			3 & 5 & 37.8 & 60.1 & 40.8 & 23.1 & 41.7 & 48.5 \\
			3 & 7 & 37.7 & 60.0 & 40.9 & 23.0 & 41.5 & 48.4 \\
			5 & 5 & 37.8 & 60.2 & 40.7 & 22.5 & 41.4 & 48.6 \\
			5 & 7 & \bf{38.1} & 60.4 & 41.3 & 23.0 & 41.6 & 48.8 \\
			7 & 7 & 38.0 & 60.2 & 41.1 & 23.0 & 41.8 & 48.8 \\ \hline
		\end{tabular}
		\vspace{-18pt}
	}
	\label{tab:k study}
	
\end{table}

Other than the softmax function, we also test other alternatives in the kernel
normalizer, such as sigmoid or sigmoid with normalization.
As shown in Table~\ref{tab:softmax}, `Softmax' and `Sigmoid Normalized'
have the same performance and better than `Sigmoid', which shows that it is
crucial to normalize the reassembly kernel to be summed to 1.

\begin{table}[t]
	\centering
	\caption{\small{Ablation study of different normalization methods in \emph{kernel normalizer}.} }
	\addtolength{\tabcolsep}{-2pt}
	\small{
		\begin{tabular}{lllllll}
			\hline
			\multirow{2}{*}{Method} & \multirow{2}{*}{AP} & \multirow{2}{*}{$\text{AP}_{50}$} & \multirow{2}{*}{$\text{AP}_{75}$} & \multirow{2}{*}{$\text{AP}_{S}$} & \multirow{2}{*}{$\text{AP}_{M}$} & \multirow{2}{*}{$\text{AP}_{L}$} \\
			&  &  &  &  &  &  \\ \hline
			Sigmoid & 37.4 & 59.8 & 40.2 & 23.1 & 40.9 & 47.4 \\
			Sigmoid Normalize & \bf{37.8} & 60.1 & 40.7 & 22.6 & 41.6 & 48.0 \\
			Softmax & \bf{37.8} & 60.1 & 40.8 & 23.1 & 41.7 & 48.5 \\ \hline
		\end{tabular}
		\vspace{-17pt}
	}
	\label{tab:softmax}
	
\end{table}

\noindent
\textbf{How \algname~Works.}
We conduct further qualitative study to figure out how \algname~works.
With a trained Mask RCNN model adopting \algname~as the upsampling operator,
we visualize the reassembling process in Figure~\ref{fig:arrow}.
In the FPN structure, the low-resolution feature map will be consecutively
upsampled for several times to a higher resolution, so a pixel in the upsampled
feature map reassembles information from a more larger region.
We sample some pixels in the high-resolution feature map, and see which neighbors it is reassembled from.
The green circle denotes example locations and red dots indicates highly weighted sources during the reassembly.
From the figure, we can clearly learn that \algname~is content-aware.
It tends to reassemble points with similar semantic information.
A location at human body prefers other points from the same human, rather than other objects or nearby background.
For locations in the background regions which has weaker semantics, the reassembly is more uniform or just biased on points with similar low-level texture features.


\section{Conclusion}
\label{sec:conclusion}

We have presented Content-Aware ReAssembly of FEatures (CARAFE), a universal, lightweight and highly effective upsampling operator. 
It consistently boosts the performances on standard benchmarks in object detection, instance/semantic segmentation and inpainting by 1.2\% AP, 1.3\% AP, 1.8\% mIoU, 1.1dB, respectively.
More importantly, \algname~introduces little computational overhead and can be readily integrated into modern network architectures. 
Future directions include exploring the applicability of \algname~in low-level vision tasks such as image restoration and super-resolution. \\
\vspace{-10pt}

\noindent\textbf{Acknowledgements.}
This work is partially supported by the Collaborative Research Grant from SenseTime Group (CUHK Agreement No. TS1610626 \& No. TS1712093), the General Research Fund (GRF) of Hong Kong (No. 14236516 \& No. 14203518), Singapore MOE AcRF Tier 1 (M4012082.020), NTU SUG, and NTU NAP.

{\small
\bibliographystyle{ieee_fullname}
\bibliography{sections/egbib}

\begin{thebibliography}{10}\itemsep=-1pt

\bibitem{chen2019hybrid}
Kai Chen, Jiangmiao Pang, Jiaqi Wang, Yu Xiong, Xiaoxiao Li, Shuyang Sun,
  Wansen Feng, Ziwei Liu, Jianping Shi, Wanli Ouyang, Chen~Change Loy, and
  Dahua Lin.
\newblock Hybrid task cascade for instance segmentation.
\newblock In {\em IEEE Conference on Computer Vision and Pattern Recognition},
  2019.

\bibitem{mmdetection}
Kai Chen, Jiaqi Wang, Jiangmiao Pang, Yuhang Cao, Yu Xiong, Xiaoxiao Li,
  Shuyang Sun, Wansen Feng, Ziwei Liu, Jiarui Xu, Zheng Zhang, Dazhi Cheng,
  Chenchen Zhu, Tianheng Cheng, Qijie Zhao, Buyu Li, Xin Lu, Rui Zhu, Yue Wu,
  Jifeng Dai, Jingdong Wang, Jianping Shi, Wanli Ouyang, Chen~Change Loy, and
  Dahua Lin.
\newblock {MMDetection}: Open mmlab detection toolbox and benchmark.
\newblock {\em arXiv preprint arXiv:1906.07155}, 2019.

\bibitem{chen2017sca}
Long Chen, Hanwang Zhang, Jun Xiao, Liqiang Nie, Jian Shao, Wei Liu, and
  Tat-Seng Chua.
\newblock {SCA-CNN}: Spatial and channel-wise attention in convolutional
  networks for image captioning.
\newblock In {\em IEEE Conference on Computer Vision and Pattern Recognition},
  2017.

\bibitem{Chen_2018_deeplab}
Liang-Chieh Chen, George Papandreou, Iasonas Kokkinos, Kevin Murphy, and Alan~L
  Yuille.
\newblock Deeplab: Semantic image segmentation with deep convolutional nets,
  atrous convolution, and fully connected crfs.
\newblock {\em IEEE Transactions on Pattern Analysis and Machine Intelligence},
  40(4):834--848, 2018.

\bibitem{Chen_2018}
Liang-Chieh Chen, Yukun Zhu, George Papandreou, Florian Schroff, and Hartwig
  Adam.
\newblock Encoder-decoder with atrous separable convolution for semantic image
  segmentation.
\newblock In {\em European Conference on Computer Vision}, 2018.

\bibitem{Dai_2017}
Jifeng Dai, Haozhi Qi, Yuwen Xiong, Yi Li, Guodong Zhang, Han Hu, and Yichen
  Wei.
\newblock Deformable convolutional networks.
\newblock In {\em IEEE International Conference on Computer Vision}, 2017.

\bibitem{Dong_2016}
Chao Dong, Chen~Change Loy, Kaiming He, and Xiaoou Tang.
\newblock Image super-resolution using deep convolutional networks.
\newblock {\em IEEE Transactions on Pattern Analysis and Machine Intelligence},
  38(2):295--307, 2016.

\bibitem{Detectron2018}
Ross Girshick, Ilija Radosavovic, Georgia Gkioxari, Piotr Doll\'{a}r, and
  Kaiming He.
\newblock Detectron.
\newblock \url{https://github.com/facebookresearch/detectron}, 2018.

\bibitem{he2017mask}
Kaiming He, Georgia Gkioxari, Piotr Doll\'{a}r, and Ross Girshick.
\newblock Mask {R-CNN}.
\newblock In {\em IEEE International Conference on Computer Vision}, 2017.

\bibitem{he2016deep}
Kaiming He, Xiangyu Zhang, Shaoqing Ren, and Jian Sun.
\newblock Deep residual learning for image recognition.
\newblock In {\em IEEE Conference on Computer Vision and Pattern Recognition},
  2016.

\bibitem{hu2019meta}
Xuecai Hu, Haoyuan Mu, Xiangyu Zhang, Zilei Wang, Tieniu Tan, and Jian Sun.
\newblock {Meta-SR}: A magnification-arbitrary network for super-resolution.
\newblock In {\em IEEE Conference on Computer Vision and Pattern Recognition},
  2019.

\bibitem{huang2019mask}
Zhaojin Huang, Lichao Huang, Yongchao Gong, Chang Huang, and Xinggang Wang.
\newblock Mask {Scoring R-CNN}.
\newblock In {\em IEEE Conference on Computer Vision and Pattern Recognition},
  2019.

\bibitem{iizuka2017globally}
Satoshi Iizuka, Edgar Simo-Serra, and Hiroshi Ishikawa.
\newblock Globally and locally consistent image completion.
\newblock {\em ACM Transactions on Graphics}, 36(4):107, 2017.

\bibitem{jaderberg2015spatial}
Max Jaderberg, Karen Simonyan, Andrew Zisserman, et~al.
\newblock Spatial transformer networks.
\newblock In {\em Advances in Neural Information Processing Systems}, 2015.

\bibitem{brab2016dynamic}
Xu Jia, Bert De~Brabandere, Tinne Tuytelaars, and Luc~V Gool.
\newblock Dynamic filter networks.
\newblock In {\em Advances in Neural Information Processing Systems}, 2016.

\bibitem{jo2018deep}
Younghyun Jo, Seoung Wug~Oh, Jaeyeon Kang, and Seon Joo~Kim.
\newblock Deep video super-resolution network using dynamic upsampling filters
  without explicit motion compensation.
\newblock In {\em IEEE Conference on Computer Vision and Pattern Recognition},
  2018.

\bibitem{kong2018deep}
Tao Kong, Fuchun Sun, Chuanqi Tan, Huaping Liu, and Wenbing Huang.
\newblock Deep feature pyramid reconfiguration for object detection.
\newblock In {\em European Conference on Computer Vision}, 2018.

\bibitem{li2016precomputed}
Chuan Li and Michael Wand.
\newblock Precomputed real-time texture synthesis with markovian generative
  adversarial networks.
\newblock In {\em European Conference on Computer Vision}, 2016.

\bibitem{li2017not}
Xiaoxiao Li, Ziwei Liu, Ping Luo, Chen~Change Loy, and Xiaoou Tang.
\newblock Not all pixels are equal: difficulty-aware semantic segmentation via
  deep layer cascade.
\newblock In {\em IEEE Conference on Computer Vision and Pattern Recognition},
  2017.

\bibitem{Lim_2017}
Bee Lim, Sanghyun Son, Heewon Kim, Seungjun Nah, and Kyoung Mu~Lee.
\newblock Enhanced deep residual networks for single image super-resolution.
\newblock In {\em IEEE Conference on Computer Vision and Pattern Recognition
  Workshop}, 2017.

\bibitem{lin2017feature}
Tsung-Yi Lin, Piotr Dollar, Ross Girshick, Kaiming He, Bharath Hariharan, and
  Serge Belongie.
\newblock Feature pyramid networks for object detection.
\newblock In {\em IEEE Conference on Computer Vision and Pattern Recognition},
  July 2017.

\bibitem{lin2014microsoft}
Tsung-Yi Lin, Michael Maire, Serge Belongie, James Hays, Pietro Perona, Deva
  Ramanan, Piotr Doll{\'a}r, and C~Lawrence Zitnick.
\newblock {Microsoft COCO}: Common objects in context.
\newblock In {\em European Conference on Computer Vision}, 2014.

\bibitem{liu2018image}
Guilin Liu, Fitsum~A Reda, Kevin~J Shih, Ting-Chun Wang, Andrew Tao, and Bryan
  Catanzaro.
\newblock Image inpainting for irregular holes using partial convolutions.
\newblock In {\em European Conference on Computer Vision}, 2018.

\bibitem{liu2018path}
Shu Liu, Lu Qi, Haifang Qin, Jianping Shi, and Jiaya Jia.
\newblock Path aggregation network for instance segmentation.
\newblock In {\em IEEE Conference on Computer Vision and Pattern Recognition},
  2018.

\bibitem{liu2015semantic}
Ziwei Liu, Xiaoxiao Li, Ping Luo, Chen-Change Loy, and Xiaoou Tang.
\newblock Semantic image segmentation via deep parsing network.
\newblock In {\em IEEE International Conference on Computer Vision}, 2015.

\bibitem{mazzini2018guided}
Davide Mazzini.
\newblock Guided upsampling network for real-time semantic segmentation.
\newblock {\em arXiv preprint arXiv:1807.07466}, 2018.

\bibitem{mildenhall2018burst}
Ben Mildenhall, Jonathan~T Barron, Jiawen Chen, Dillon Sharlet, Ren Ng, and
  Robert Carroll.
\newblock Burst denoising with kernel prediction networks.
\newblock In {\em IEEE Conference on Computer Vision and Pattern Recognition},
  2018.

\bibitem{molchanov2016pruning}
Pavlo Molchanov, Stephen Tyree, Tero Karras, Timo Aila, and Jan Kautz.
\newblock Pruning convolutional neural networks for resource efficient
  inference.
\newblock {\em arXiv preprint arXiv:1611.06440}, 2016.

\bibitem{Newell_2016}
Alejandro Newell, Kaiyu Yang, and Jia Deng.
\newblock Stacked hourglass networks for human pose estimation.
\newblock In {\em European Conference on Computer Vision}, 2016.

\bibitem{Noh_2015}
Hyeonwoo Noh, Seunghoon Hong, and Bohyung Han.
\newblock Learning deconvolution network for semantic segmentation.
\newblock {\em IEEE International Conference on Computer Vision}, Dec 2015.

\bibitem{pang2019libra}
Jiangmiao Pang, Kai Chen, Jianping Shi, Huajun Feng, Wanli Ouyang, and Dahua
  Lin.
\newblock Libra {R-CNN}: Towards balanced learning for object detection.
\newblock In {\em IEEE Conference on Computer Vision and Pattern Recognition},
  2019.

\bibitem{pathak2016context}
Deepak Pathak, Philipp Krahenbuhl, Jeff Donahue, Trevor Darrell, and Alexei~A
  Efros.
\newblock Context encoders: Feature learning by inpainting.
\newblock In {\em IEEE Conference on Computer Vision and Pattern Recognition},
  2016.

\bibitem{ren2015faster}
Shaoqing Ren, Kaiming He, Ross Girshick, and Jian Sun.
\newblock Faster {R-CNN}: Towards real-time object detection with region
  proposal networks.
\newblock In {\em Advances in Neural Information Processing Systems}, 2015.

\bibitem{ronneberger2015u}
Olaf Ronneberger, Philipp Fischer, and Thomas Brox.
\newblock U-net: Convolutional networks for biomedical image segmentation.
\newblock In {\em International Conference on Medical image computing and
  computer-assisted intervention}, 2015.

\bibitem{Shi_2016}
Wenzhe Shi, Jose Caballero, Ferenc Husz{\'a}r, Johannes Totz, Andrew~P Aitken,
  Rob Bishop, Daniel Rueckert, and Zehan Wang.
\newblock Real-time single image and video super-resolution using an efficient
  sub-pixel convolutional neural network.
\newblock In {\em IEEE Conference on Computer Vision and Pattern Recognition},
  2016.

\bibitem{ulyanov2018deep}
Dmitry Ulyanov, Andrea Vedaldi, and Victor Lempitsky.
\newblock Deep image prior.
\newblock In {\em IEEE Conference on Computer Vision and Pattern Recognition},
  2018.

\bibitem{wang2019region}
Jiaqi Wang, Kai Chen, Shuo Yang, Chen~Change Loy, and Dahua Lin.
\newblock Region proposal by guided anchoring.
\newblock In {\em IEEE Conference on Computer Vision and Pattern Recognition},
  2019.

\bibitem{xiao2018unified}
Tete Xiao, Yingcheng Liu, Bolei Zhou, Yuning Jiang, and Jian Sun.
\newblock Unified perceptual parsing for scene understanding.
\newblock In {\em European Conference on Computer Vision}, 2018.

\bibitem{Xiong_2019_CVPR}
Wei Xiong, Jiahui Yu, Zhe Lin, Jimei Yang, Xin Lu, Connelly Barnes, and Jiebo
  Luo.
\newblock Foreground-aware image inpainting.
\newblock In {\em IEEE Conference on Computer Vision and Pattern Recognition},
  June 2019.

\bibitem{Xu_2019_CVPR}
Rui Xu, Xiaoxiao Li, Bolei Zhou, and Chen~Change Loy.
\newblock Deep flow-guided video inpainting.
\newblock In {\em IEEE Conference on Computer Vision and Pattern Recognition},
  June 2019.

\bibitem{yu2018free}
Jiahui Yu, Zhe Lin, Jimei Yang, Xiaohui Shen, Xin Lu, and Thomas~S Huang.
\newblock Free-form image inpainting with gated convolution.
\newblock {\em arXiv preprint arXiv:1806.03589}, 2018.

\bibitem{yu2018generative}
Jiahui Yu, Zhe Lin, Jimei Yang, Xiaohui Shen, Xin Lu, and Thomas~S Huang.
\newblock Generative image inpainting with contextual attention.
\newblock In {\em IEEE Conference on Computer Vision and Pattern Recognition},
  2018.

\bibitem{zhao2017pyramid}
Hengshuang Zhao, Jianping Shi, Xiaojuan Qi, Xiaogang Wang, and Jiaya Jia.
\newblock Pyramid scene parsing network.
\newblock In {\em IEEE Conference on Computer Vision and Pattern Recognition},
  2017.

\bibitem{zhao2018psanet}
Hengshuang Zhao, Yi Zhang, Shu Liu, Jianping Shi, Chen Change~Loy, Dahua Lin,
  and Jiaya Jia.
\newblock {PSANet}: Point-wise spatial attention network for scene parsing.
\newblock In {\em European Conference on Computer Vision}, 2018.

\bibitem{zhao2019m2det}
Qijie Zhao, Tao Sheng, Yongtao Wang, Zhi Tang, Ying Chen, Ling Cai, and Haibin
  Ling.
\newblock {M2Det}: A single-shot object detector based on multi-level feature
  pyramid network.
\newblock In {\em AAAI Conference on Artificial Intelligence}, 2019.

\bibitem{zhou2017places}
Bolei Zhou, Agata Lapedriza, Aditya Khosla, Aude Oliva, and Antonio Torralba.
\newblock Places: A 10 million image database for scene recognition.
\newblock {\em IEEE Transactions on Pattern Analysis and Machine Intelligence},
  2017.

\bibitem{zhou2017scene}
Bolei Zhou, Hang Zhao, Xavier Puig, Sanja Fidler, Adela Barriuso, and Antonio
  Torralba.
\newblock Scene parsing through {ADE20K} dataset.
\newblock In {\em IEEE Conference on Computer Vision and Pattern Recognition},
  2017.

\bibitem{zhou2018semantic}
Bolei Zhou, Hang Zhao, Xavier Puig, Tete Xiao, Sanja Fidler, Adela Barriuso,
  and Antonio Torralba.
\newblock Semantic understanding of scenes through the ade20k dataset.
\newblock {\em International Journal of Computer Vision}, 2018.

\end{thebibliography}
}
\clearpage

\appendix
\begin{appendices}
\section{Detail Experimental Settings}
\label{sec:details}

\noindent\textbf{Object Detection and Instance Segmentation.}
\label{subsec:detection}
We evaluate \algname~on Faster RCNN~\cite{ren2015faster} and Mask RCNN~\cite{he2017mask} with the ResNet-50 backbone~\cite{he2016deep}.
FPN~\cite{lin2017feature} is used for these methods.
In both training and inference, we resize an input image such that its shorter edge has 800 pixels or longer edge has 1333 pixels without changing its aspect ratio.
We adopt synchronized SGD with an initial learning rate of 0.02, a momentum of 0.9 and a weight decay of 0.0001. We use a batchsize of 16 over 8 GPUs 
(2 images per GPU). Following the 1x training schedule as Detectron~\cite{Detectron2018} and MMDetection~\cite{mmdetection}, 
we train 12 epochs in total and decrease the learning rate by a factor of 0.1 at epoch 8 and 11.

\noindent\textbf{Semantic Segmentation.}
\label{subsec:segmentation}
We use the official implementation of UperNet\footnote{\href{https://github.com/CSAILVision/semantic-segmentation-pytorch}{https://github.com/CSAILVision/semantic-segmentation-pytorch}}~\cite{xiao2018unified} with the 
ResNet-50 backbone. During the training, an input image is resized such that the size of its shorter edge is randomly selected from \{300, 375, 450, 525, 600\}. In inference, we apply the single scale 
testing for a fair comparison and the shorter edge of an image is set to 450 pixels. The maximum length of the longer edge of an image is set to 1200 in both training and inference. 
We adopt synchronized SGD with an initial learning rate of 0.02, a momentum of 0.9 and a weight decay of 0.0001. We use a batchsize of 16 over 8 GPUs 
(2 images per GPU), and synchronized batch normalization is adopted as a common practice in semantic segmentation. Following~\cite{Chen_2018_deeplab}, the `poly' learning rate policy in which the 
learning rate of current iteration equals to the initial learning rate multiplying $(1-iter/max\_iter)^{power}$ is adopted. We set $power$ to 0.9 and train 20 epochs in total.

\noindent\textbf{Image Inpainting.}
\label{subsec:inpainting}
We employ the generator and discriminator networks from Global\&Local~\cite{iizuka2017globally} as the baseline. 
Our generator takes a $256 \times 256$ image $\mathbf{x}$ with masked region $M$ as input and produces a $256 \times 256$ prediction of the missing region $\mathbf{\hat{y}}$ as output. 
Then we combine the predicted image with the input by $\mathbf{y}=(1-M)\odot \mathbf{x} + M \odot \mathbf{\hat{y}}$.
Finally, the combined output $\mathbf{y}$ is fed into the discriminator.
We apply a simple modification to the baseline model to achieve better generation quality.
Compared to the original model that employs two discriminators, we employ only one PatchGAN-style discriminator\cite{li2016precomputed} on the inpainted region. 
This modification can achieve better image quality.

For a fair comparison and taking real-world application into consideration, we use the free-form masks introduced by \cite{yu2018free} as the binary mask $M$. 
For Partial Conv~\cite{liu2018image}, we just substitute the convolution layers with the official Partial Conv module in our generator.
During training, Adam solver with learning rate 0.0001 is adopted where $\beta_1=0.5$ and $\beta_2=0.9$. Training batch size is 32.
The input and output are linearly scaled within range $[-1, 1]$. 

\section{Visualization of \algname~}
\label{sec:qualitative}

\begin{figure*}[ht]
	\centering
	\includegraphics[width=\linewidth]{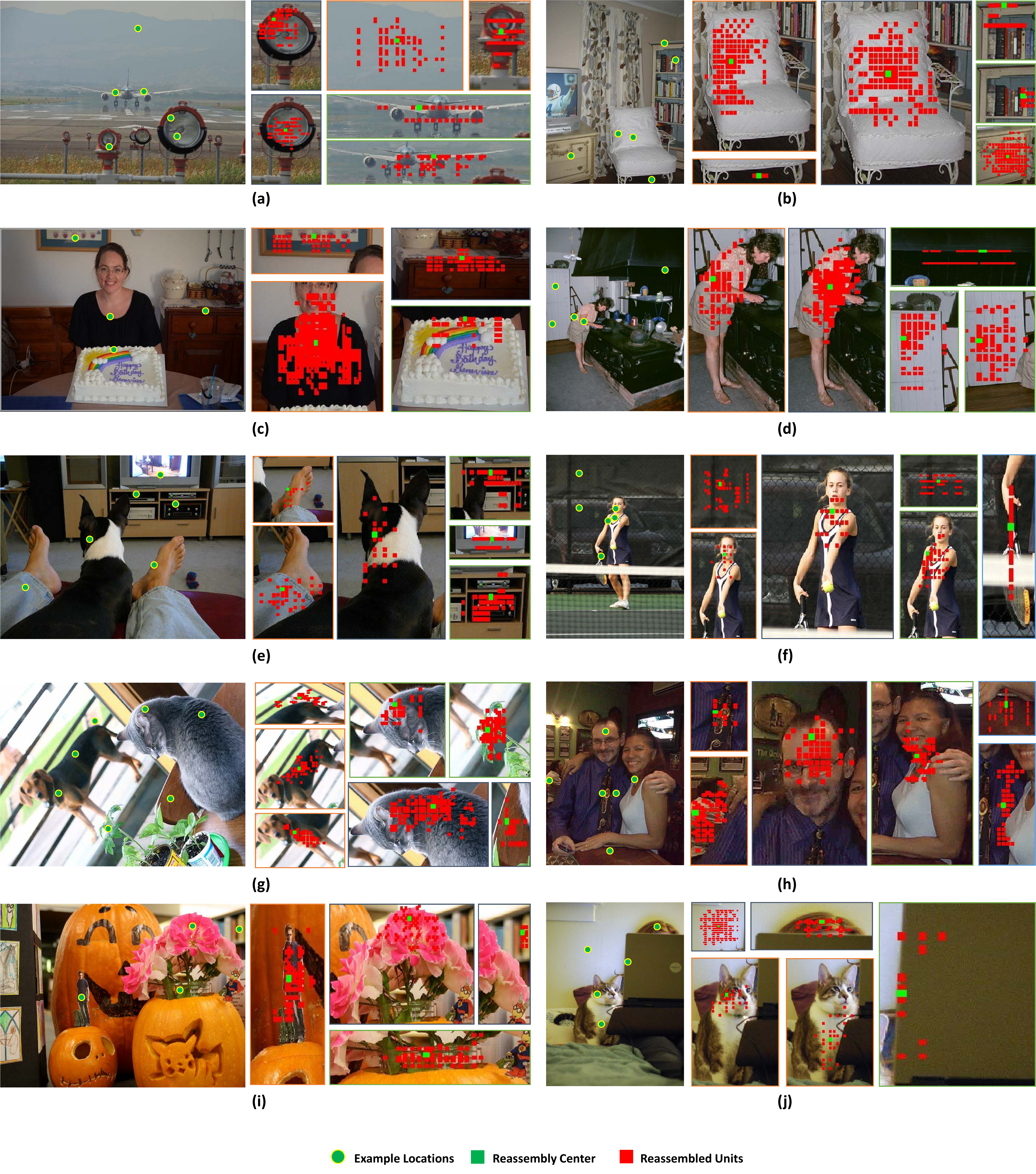}
	\caption{\small{\algname~performs content-aware reassembly when upsampling a feature map. 
			Red units are reassembled into the green center unit by \algname~in the top-down pathway of a FPN structure. }}
	\label{fig:qualitive}
\end{figure*}

We demonstrate how \algname~performs content-aware reassembly with more examples in Figure~\ref{fig:qualitive}. 
Red units are reassembled into the green center unit by \algname~in the top-down pathway of a FPN structure.

\section{Visual Results Comparison}
\begin{figure*}[ht]
	\centering
	\includegraphics[width=\linewidth]{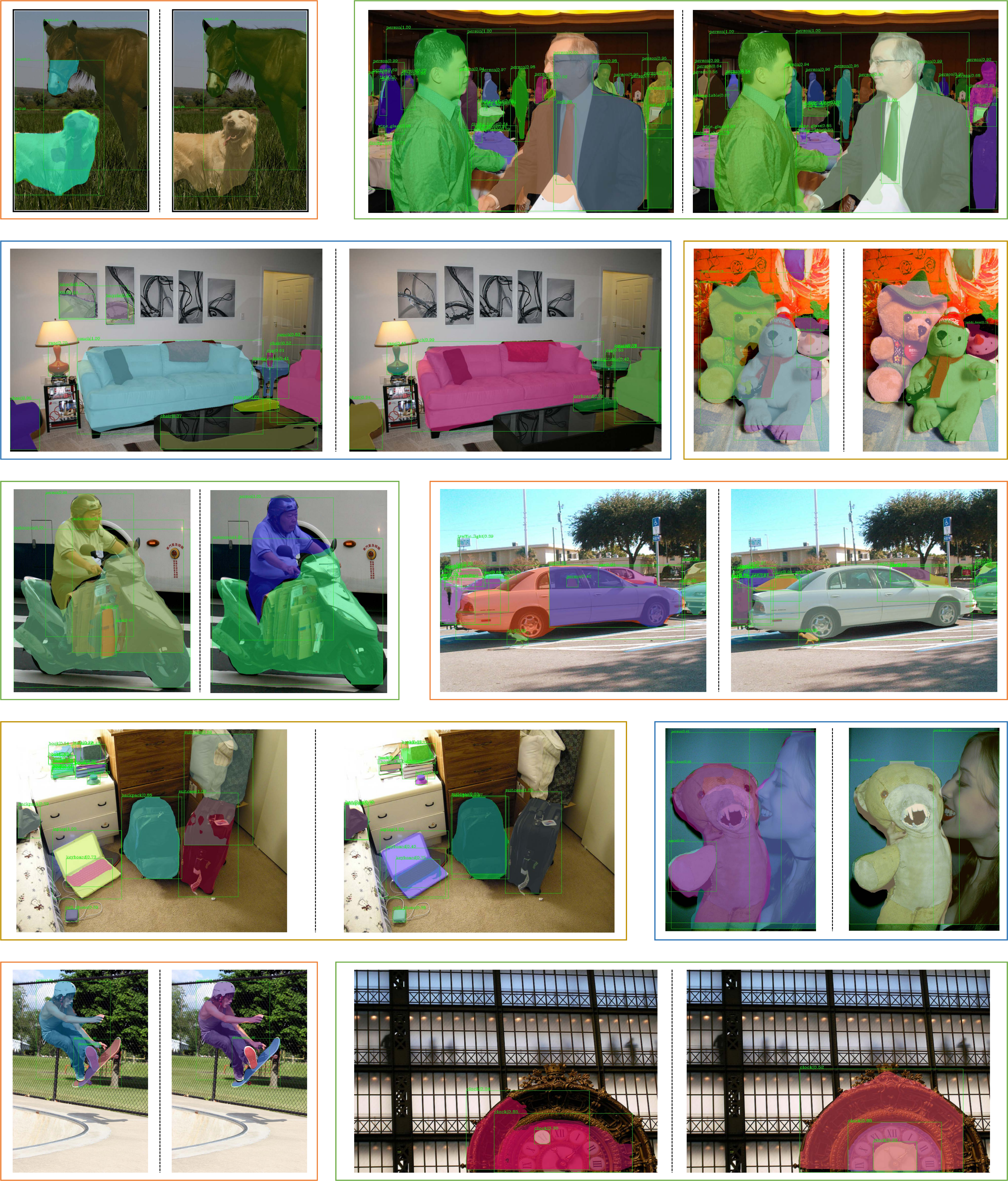}
	\caption{\small{More comparison of object detection and instance segmentation results between Mask RCNN~\cite{he2017mask} baseline (left to the dash line) and Mask RCNN w/ \algname~ (right to the dash line) on COCO 2017 val.}}
	\vspace{-10pt}
	\label{fig:detection}
\end{figure*}

\noindent\textbf{Object Detection and Instance Segmentation.}
As illustrated in Figure~\ref{fig:detection}, we provide more object detection and instance segmentation results comparison between Mask RCNN baseline and Mask RCNN w/ \algname~on COCO~\cite{lin2014microsoft} 2017 val.

\begin{figure*}[ht]
	\centering
	\includegraphics[width=\linewidth]{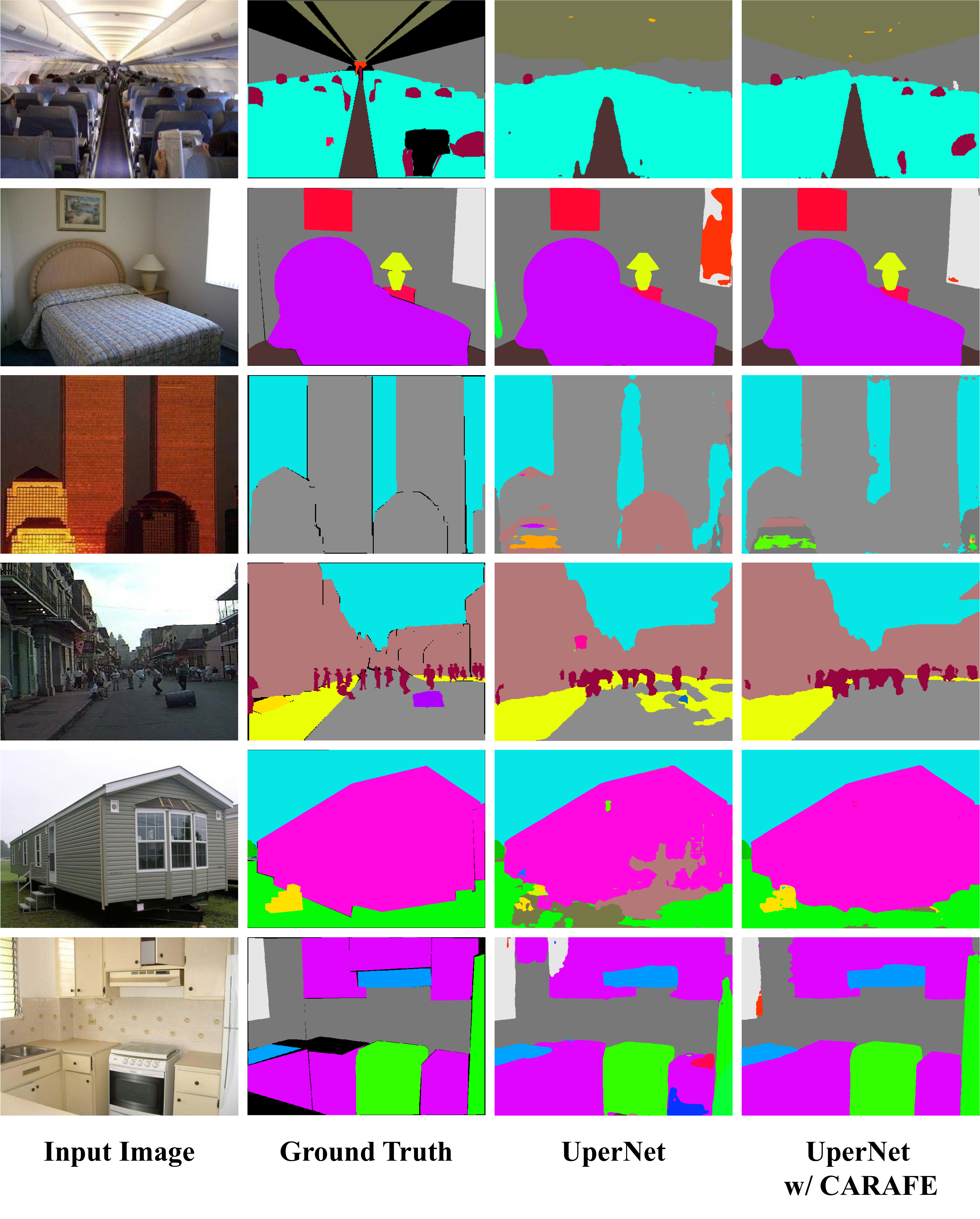}
	\caption{\small{Comparison of semantic segmentation results between UperNet~\cite{xiao2018unified} baseline and UperNet w/ \algname~on ADE20k val. Columns from left to right correspond to the input image, ground truth, baseline results and \algname~results, respectively.}}
	\vspace{-10pt}
	\label{fig:semantic}
\end{figure*}

\noindent\textbf{Semantic Segmentation.}
We compare the semantic segmentation results between UperNet baseline and UperNet w/ \algname~on ADE20k~\cite{zhou2017scene} val in Figure~\ref{fig:semantic}.

\begin{figure*}[ht]
	\centering
	\includegraphics[width=0.9\linewidth]{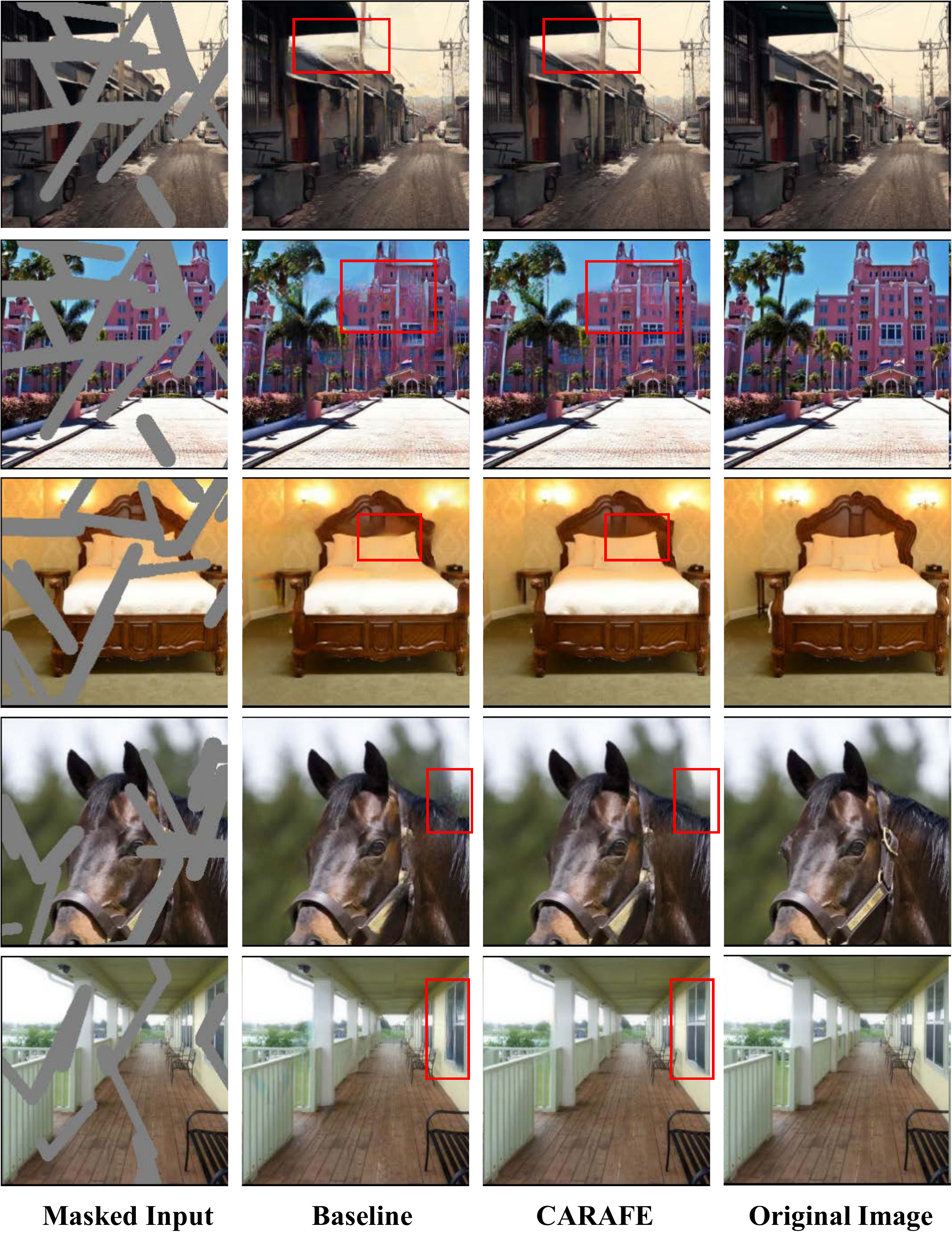}
	\caption{\small{Comparison of image inpainting results between Global\&Local~\cite{iizuka2017globally} baseline and Global\&Local w/ \algname~on Places val. Columns from left to right correspond to the masked input, baseline results, \algname~results and original image, respectively.}}
	\vspace{-10pt}
	\label{fig:inpaint}
\end{figure*}

\noindent\textbf{Image Inpainting.}
Comparison of image inpainting results between Global\&Local baseline and Global\&Local w/ \algname~on Places\cite{zhou2017places} val is shown in Figure~\ref{fig:inpaint}.
\end{appendices}

\clearpage
\clearpage
\clearpage
\clearpage

\end{document}